\documentclass[a4paper,twoside]{article}

\usepackage{epsfig}
\usepackage{subcaption}
\usepackage{calc}
\usepackage{amssymb}
\usepackage{amstext}
\usepackage{amsmath}
\usepackage{amsthm}
\usepackage{multicol}
\usepackage{pslatex}
\usepackage{apalike}
\usepackage{algorithm2e}
\usepackage[bottom]{footmisc}
\usepackage{url}
\usepackage{xcolor}

\usepackage{ragged2e}
\usepackage{booktabs}
\usepackage{tabularx}

\usepackage{SCITEPRESS}     % Please add other packages that you may need BEFORE the SCITEPRESS.sty package.

\begin{document}

\title{Contrato\textit{\textcolor{red}{3}\textcolor{green}{6}\textcolor{blue}{0}} 2.0: A Document and Database-Driven Question-Answer System using Large Language Models and Agents}

\author{
\authorname{Antony Seabra\sup{1,2}
Claudio Cavalcante\sup{1,2}
João Nepomuceno\sup{1} \\
Lucas Lago\sup{1}
Nicolaas Ruberg\sup{1} and 
Sergio Lifschitz\sup{2}}
\affiliation{\sup{1}BNDES - Área de Tecnologia da Informação, Rio de Janeiro, Brazil}
\affiliation{\sup{2}PUC-Rio - Departamento de Informática, Rio de Janeiro, Brazil}
}


\abstract{We present a question-and-answer (Q\&A) application designed to support the contract management process by leveraging combined information from contract documents (PDFs) and data retrieved from contract management systems (database). This data is processed by a large language model (LLM) to provide precise and relevant answers. The accuracy of these responses is further enhanced through the use of Retrieval-Augmented Generation (RAG), text-to-SQL techniques, and agents that dynamically orchestrate the workflow. These techniques eliminate the need to retrain the language model. Additionally, we employed Prompt Engineering to fine-tune the focus of responses. Our findings demonstrate that this multi-agent orchestration and combination of techniques significantly improve the relevance and accuracy of the answers, offering a promising direction for future information systems.}

\onecolumn \maketitle \normalsize \setcounter{footnote}{0} \vfill

\section{\uppercase{Introduction}}
\label{sec:introduction}
Contract management in large corporations involves overseeing legally binding agreements from their initiation through to execution and finalization. This process encompasses ensuring that services or products are delivered in accordance with contractual terms, monitoring their execution, and continuously evaluating both operational and financial performance throughout the service or product lifecycle. In the case of public sector companies, this process becomes even more complex due to stringent regulatory frameworks. In Brazil, for instance, Law No. 14,133/2021 mandates that contract management includes a wide range of activities, such as technical and administrative oversight, adherence to contract duration, re-evaluation of economic and financial terms, modifications to service scope, and the enforcement of penalties and fines when necessary. These regulations impose an additional layer of complexity on the contract management process, demanding a robust and systematic approach to ensure compliance and efficiency.

Beyond contract managers, dedicated organizational units are essential to support the contract management process, ensuring that the diverse range of activities associated with contract execution is managed efficiently. Often, these units require specialized knowledge to handle complex services effectively. Notable examples include information and communication technology (ICT) services, property and asset management, and construction and engineering projects, each of which demands a high level of expertise. Additionally, these units typically rely on Contract Management Systems (CMS) to streamline their operations. Public companies may either develop these systems in-house or opt for widely-used market solutions, such as SAP Contract Life-cycle Management and IBM Emptoris Contract Management, among others.

While these systems efficiently handle general contract information, such as signatures, expiration dates, payment terms, and contract agents, many specific details required to support effective management activities remain accessible only through the original documents. For instance, traditional Contract Management Systems (CMS) are often unable to respond to inquiries concerning particular aspects of a contract, such as penalties, discounts, or fines associated with delays in service or product delivery. Moreover, they lack the capability to provide insights into comparative characteristics across different contracts, such as penalty clauses related to database support agreements. These tasks are highly time-consuming.

The objective of this study is to provide a solution that aids contract managers in addressing queries related to both contract documents and data housed within traditional Contract Management Systems. One of the key challenges faced by contract managers is the time-consuming process of searching for and retrieving relevant information from lengthy and complex contract texts. To address this, we leverage state-of-the-art large-scale Language Modeling (LLM) technologies to analyze and extract pertinent details from contract documents efficiently. This not only improves the accuracy of the information retrieved but also significantly enhances the productivity of contract managers by reducing the manual effort required to locate specific information. Additionally, our approach integrates data from traditional Contract Management Systems, ensuring that responses are both relevant and comprehensive, thereby streamlining contract management activities.

In this work, we evaluated and integrated several Natural Language Processing (NLP) techniques to develop a Q\&A system specifically designed for IC contracts, using contract PDF files and data from Contract Management Systems (CMS) as primary data sources. To enhance the relevance of user queries, prior work by \cite{sbbdContrato360} employed Retrieval-Augmented Generation (RAG) techniques and a static approach to text-to-SQL for extracting relevant metadata from contract systems. Building upon this, our approach utilizes agents to dynamically improve the accuracy and contextual relevance of responses, with a particular focus on a context-aware text-to-SQL agent that interprets user queries more effectively. Furthermore, similar to \cite{sbbdContrato360}, we applied Prompt Engineering techniques to standardize responses and ensure greater precision in the answers provided.

One of the primary challenges in interpreting contract documents lies in distinguishing between relevance and similarity, a complexity that arises due to the standardized formats and repetitive textual structures commonly found in these documents. This standardization is a challenge for LLMs because there is a great deal of textual similarity, which does not necessarily translate into relevance. Using a mix of NLP techniques, we developed a solution that minimizes the impact of standardization and provides relevant answers. This approach made it possible to design a solution without needing traditional \emph{fine-tuning} or re-training of language models.

The paper is organized as follows: Section 2 provides technical background on LLMs, \textit{RAGs} \textit{text-to-SQL}, \textit{agents}, and \textit{prompt} engineering. Section 3 discusses the methodology of the use of the presented techniques, while Section 4 details the architecture of our solution. Section 5 describes how we evaluated the proposed solution and the experimentation of the Q\&A application. Finally, Section 6 concludes our study and proposes directions for future research in this field.

\section{\uppercase{Background}}
The dissemination of several applications in the area of Natural Language Processing (NLP) was made possible by Large Scale Language Models (LLMs), including question and answer (Q\&A) systems. Recently, the use of agents has been introduced as a crucial component in LLM-based systems to orchestrate and manage task execution dynamically. Agents, such as router agents, SQL agents, and RAG agents, enable the efficient allocation of tasks by directing queries to the most suitable processing modules, enhancing system adaptability and performance. This approach allows LLMs to better handle complex queries, making responses more accurate and contextually relevant by integrating external data sources and specialized processing routines \cite{mialon2023augmented}.

\subsection{Large Language Models}
Large-scale Language Models (LLMs) have revolutionized the field of natural language processing with their ability to understand and generate human-like text. In their architecture, they utilize a specific neural network structure, \textit{Transformers}, which allows the model to weight the influence of different parts of the input texts at different times \cite{vaswani2017attention}.

Conversational applications, a specific use case for LLMs, specialize in generating text that is coherent and contextualized. This is achieved through training, in which the models are fed vast amounts of conversational data, allowing them to learn the nuances of dialogue \cite{OpenAI}. In this way, LLMs have established a new paradigm for NLP. Moreover, by expanding the search space with external data or specializing through fine-tuning, LLMs become platforms for building specialized applications. In this work, all language models utilized were based on OpenAI's GPT series. Specifically, we employed the \textit{text-davinci-002} model for generating embeddings and the \textit{gpt-4-turbo} model for generating answers to user queries.

\subsection{Retrieval-Augmented Generation (RAG)}
According to \cite{chen2024benchmarking}, LLMs face significant challenges such as factual hallucination, outdated knowledge, and lack of domain-specific \textit{expertise}. In response to these challenges, RAG represents a paradigm shift in the way LLMs process and generate text. The principle behind RAG involves using vector storage to retrieve text fragments similar to the input query \cite{gao2023retrieval}. This technique converts both the query text and the information database into high-dimensional vectors, allowing one to retrieve similar information, which is then fed to an LLM.

\begin{figure*}[ht]
\centering
\includegraphics[width=\textwidth]{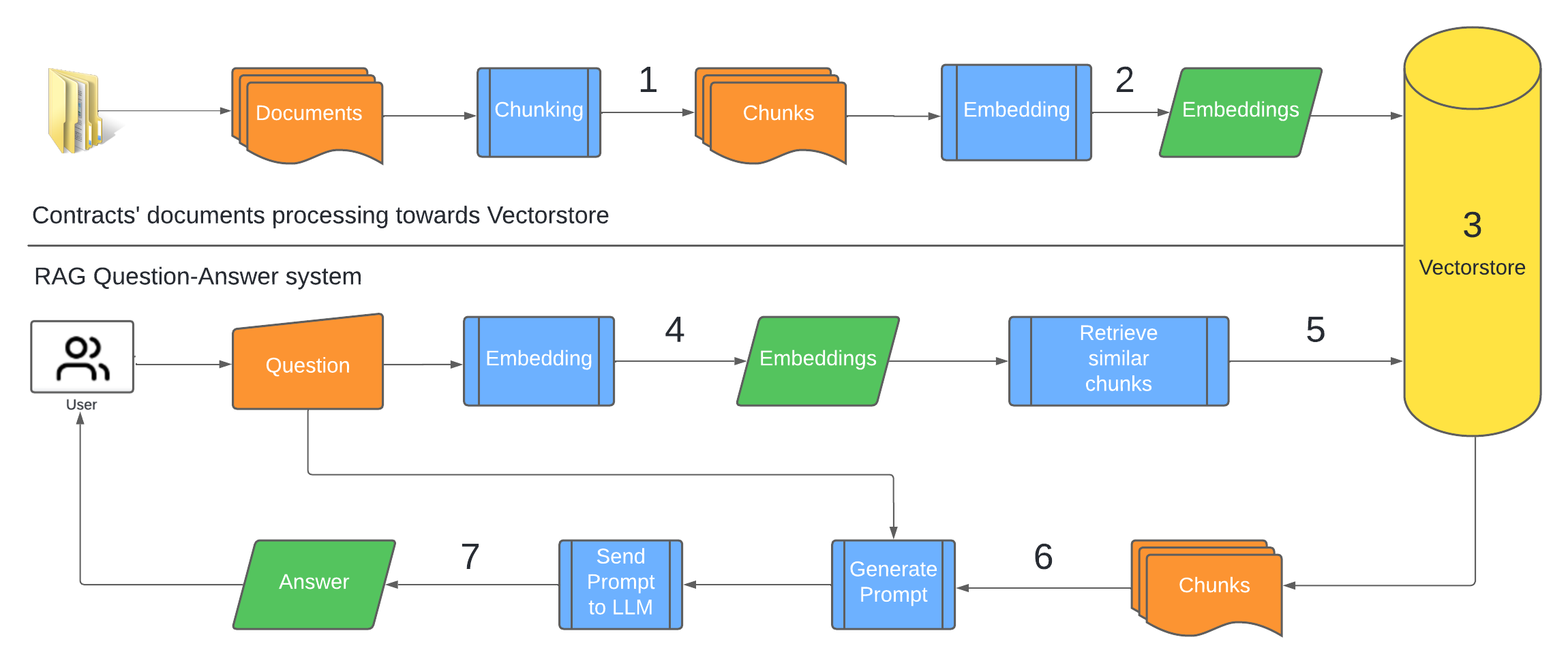}
\caption{Retrieval-Augmented Generation}
\label{fig:rag}
\end{figure*}

\cite{gao2023retrieval} and \cite{feng2024retrieval} describe \textit{frameworks} that exploit the advantages of this technique by providing additional data to the LLM without re-training the \cite{li2022survey} model. By dividing the available text into manageable chunks and embedding these chunks in high-dimensional vector spaces, it is possible to quickly retrieve contextually relevant information in response to a query, which informs the next processing steps. As shown in Figure \ref{fig:rag}, the first step (1) involves reading the textual content of the PDF documents into manageable chunks (\textit{chunks}), which are then transformed (\textit{embedding}) (2) into high-dimensional vectors. The text in vector format captures the semantic properties of the text, a format that can have 1536 dimensions.

These \textit{embeddings} vectors are stored in a \textit{vectorstore} (3), a database specialized in high-dimensional vectors. The vector store allows efficient querying of vectors through their similarities, using the distance for comparison (whether \textit{Manhatan}, Euclidean or cosine).

Once the similarity metric is established, the query is \textit{embedded} in the same vector space (4); this allows a direct comparison between the vectorized query and the vectors of the stored chunks, retrieving the most similar chunks (5), which are then transparently integrated into the LLM context to generate a \textit{prompt} (6). The \textit{prompt} is then composed of the question, the texts retrieved from the \textit{vectorstore}, the specific instructions and, optionally, the \textit{chat} history, all sent to the LLM which generates the final response (7).

In RAG, the \textit{chunking} strategy is important because it directly influences the quality of the retrieved information. A well-designed chunk generation ensures that the information is cohesive and semantically complete, capturing its essence.

A key aspect of RAG is the difference between similarity and relevance. Similar passages may not contain the information relevant to answering a query, posing a challenge to accurately retrieve information, especially in cases where data comes from multiple documents with similar structure. In such contexts, documents may share a high degree of structural and lexical similarity, making it difficult for retrieval algorithms to distinguish between content that is merely similar in form and content that is truly relevant to a query.

\subsection{Text-to-SQL}
\textit{Text-to-SQL} is a technology that enables the conversion of natural language queries into SQL commands based solely on the database schema, eliminating the need for knowledge of the underlying data \cite{liu2023comprehensive}. This approach leverages the capabilities of LLMs to understand and interpret human language, allowing users to retrieve data from databases through plain text input without requiring specialized knowledge of SQL syntax \cite{gao2023text}.

By translating natural language into SQL queries, \textit{text-to-SQL} brings complex database structures and end users closer together, making access more intuitive and efficient. This technique is particularly useful because it allows non-expert users to access databases by asking natural language queries. It improves data accessibility, reduces the learning curve associated with database querying, and speeds up data analysis processes, enabling more users to make data-driven decisions.

The main distinction between RAG and \textit{text-to-SQL} techniques lies in how information is retrieved. RAG relies on retrieving text segments from a vector store that are similar to the user's question, and using these segments to generate a coherent and contextually relevant answer. This method is effective for questions where the answer can be synthesized from existing text. However, it is not always possible to identify the information expected as the answer. In another aspect, \textit{text-to-SQL} translates natural language queries into SQL commands, as demonstrated in \cite{pinheiro2023construction}, which are then executed against a structured database to retrieve exact data matches. This ensures that if the text-to-SQL translation is accurate, the user will receive a highly specific answer directly from the database fields.

Therefore, while RAG operates on the principle of textual similarity and generative capabilities, \textit{text-to-SQL} offers a more intrusive mechanism for data retrieval by executing queries that directly match the user's intent, making it particularly effective for data investigations.

\subsection{Prompt Engineering}
Prompt engineering is the art of designing and optimizing \textit{prompts} to guide LLMs in generating desired outputs. The goal of \textit{prompt} engineering is to maximize the potential of LLMs by providing them with instructions and context \cite{OpenAIprompt}.

In the context of \textit{prompt} engineering, prompts are a fundamental part of the process. Through prompts, engineers can outline the script for a response, specifying the desired style and format for the LLM response \cite{white2023prompt} \cite{giray2023prompt}. For example, to define the style of a conversation, a \textit{prompt} could be formulated as "Use professional language and treat the customer with respect" or "Use informal language and emojis to convey a friendly tone." To specify the format of dates in responses, a \textit{prompt} instruction could be "Use the American format, MM/DD/YYYY, for all dates."

On the other hand, context refers to the information provided to LLMs along with the main prompts. The most important aspect of context is that it can provide additional information to support the response given by the LLM, which is very useful when implementing Q\&A systems. This supplemental context can include relevant background details, specific examples, and even previous dialogue exchanges, which collectively help the model generate more accurate, detailed, and contextually appropriate responses. According to \cite{wang2023unleashing}, \textit{prompts} provide guidance to ensure that the model generates responses that are aligned with the user’s intent. As a result, well-crafted \textit{prompts} significantly improve the effectiveness and appropriateness of responses.

Recent studies have begun to explore the synergistic integration of these techniques with LLMs to create more sophisticated Q\&A systems. For example, \cite{jeong2023study} reinforces the importance of using Prompt Engineering with RAG to improve the retrieval of relevant documents, which are then used to generate both contextually relevant and information-rich answers. Similarly, \cite{gao2023text} explores the integration of \textit{text-to-SQL} with Prompt Engineering to enhance the model's ability to interact directly with relational databases, thereby expanding the scope of queries that can be answered accurately.

\subsection{Agents}
The use of agents in applications built around Large Language Models (LLMs) is relatively recent but has already became common. Agents act as intelligent intermediaries that route, process, and present information in ways tailored to the context of the query. These agents leverage recent advancements in AI, such as Retrieval-Augmented Generation (RAG)
and tool utilization, to perform more complex and contextually aware tasks \cite{lewis2020retrieval}. They play a pivotal role in orchestrating complex tasks, integrating various data sources, and ensuring that the system responds accurately and efficiently to user queries.

In a complex LLM-based system, different tasks often require specialized handling. Agents enable task orchestration by directing queries to the most appropriate component, whether it's for retrieving data, performing calculations, or generating visualizations. For example, an application may have a Text-to-SQL agent to perform queries over a relational database and a Graph agent to visualize graphs after an answer, if appropriate. According to \cite{jin2024llms}, applying LLMs to text-to-database management and query optimization is also a novel research direction in natural language to code generation task. By converting natural language queries into SQL statements, LLMs help developers quickly generate efficient database query code. In the realm of integrating heterogeneous data sources, Q\&A applications often need to access data from documents, databases, APIs, and other repositories. Agents facilitate the seamless integration of these heterogeneous data sources, allowing the system to extract relevant information dynamically.

There are several agent types. As outlined in \cite{singh2024enhancing}, agent workflows allow LLMs to operate more dynamically by incorporating specialized agents that manage task routing, execution, and optimization. These agents serve as intelligent intermediaries, directing specific tasks—such as data retrieval, reasoning, or response generation—to the most suitable components within the system. One of the most important ones in place are the Router Agents, as they are the decision-makers of the system. When a user poses a query, the router agent analyzes the input and decides the best path forward. For instance, if a query is identified as needing factual data, the router agent might direct it to a RAG model. If the question involves specific data retrieval from a database, it will engage an SQL agent instead.

As mentioned before, RAG and SQL Agents are very relevant too. According to \cite{saeed2023querying}, SQL agents can effectively manage data retrieval tasks by leveraging LLMs. The SQL queries are transformed into prompts for LLMs, allowing the system to interact with unstructured data stored in the model, mimicking traditional database operations. \cite{fan2024survey} provides a comprehensive overview of the integration of RAG techniques in LLMs but moreover, \cite{wang2024speculative} introduces a novel approach that combines RAG techniques with a drafting-verification process to improve the reasoning capabilities of LLMs when handling retrieved documents. The RAG agent, termed the "drafter," generates multiple answer drafts based on retrieved results, while a larger generalist LLM, the "verifier," assesses these drafts and selects the most accurate one. This approach effectively integrates retrieval and generation, enhancing the overall performance of LLMs in knowledge-intensive tasks such as question answering and information retrieval systems.

\section{\uppercase{Methodology}}
To address the challenges faced by contract managers in terms of complex information retrieval, we propose Contrato360, a Q\&A system supported by an LLM and orchestrated by agents. The system employs a range of techniques designed to enhance the relevance of responses while mitigating the risks associated with the standardized textual structures of contracts.

\begin{figure*}[ht]
\centering
\includegraphics[width=1\textwidth]
{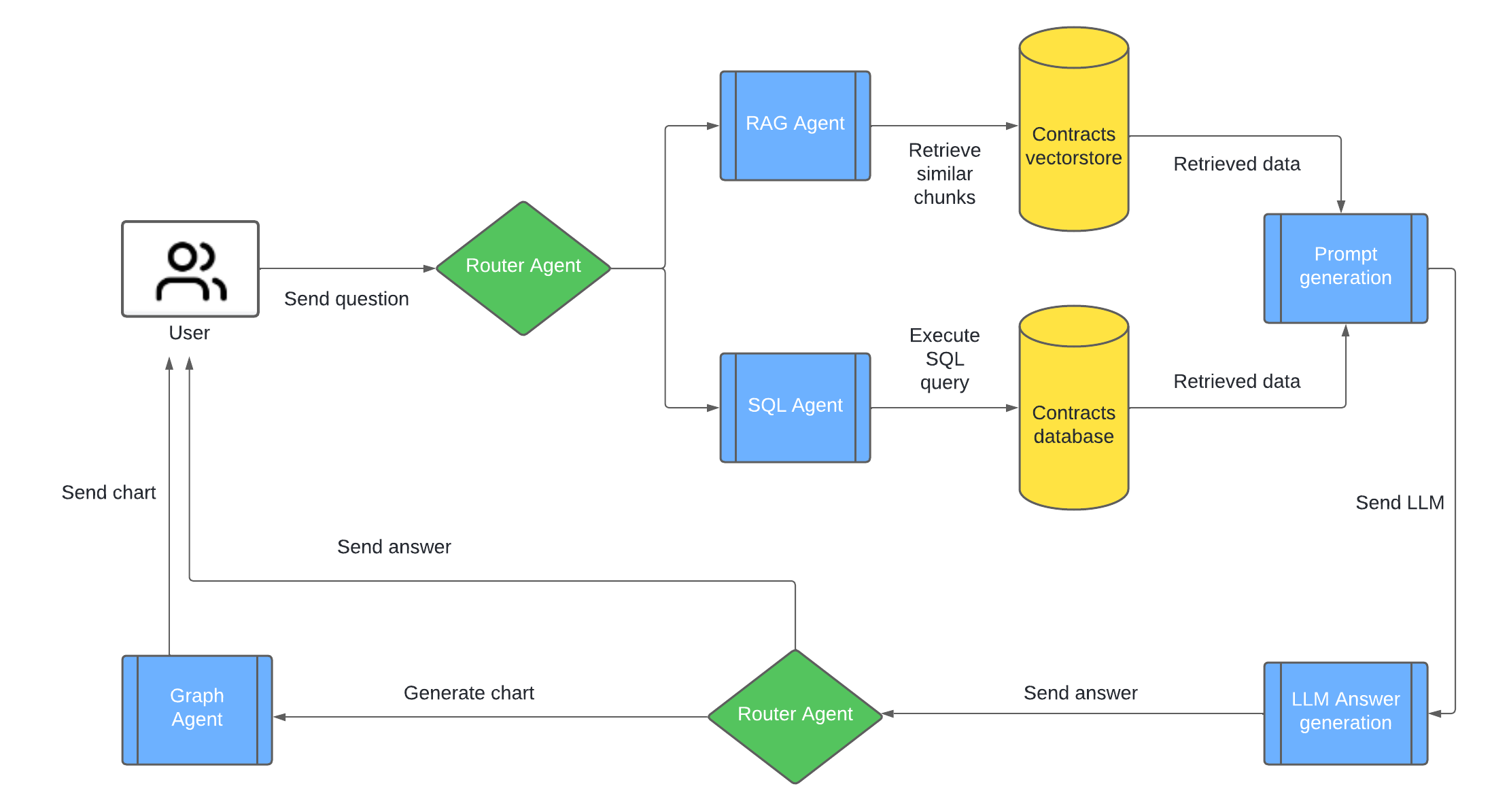}
\caption{Methodology Workflow Combining Different Techniques}
\label{fig:agents}
\end{figure*}

To achieve this goal of increasing the relevance of the responses obtained by Contrato\textit{360}, we combined four techniques: 1) Retrieval-Augmented Generation (RAG) to increase the relevance of information about contracts contained in PDF documents; 2) Agents to orchestrate and route the flow of execution, enabling the dynamic selection of the most appropriate approach for each query context; 3) Text-to-SQL agent to retrieve the relevant information from contract systems; 4) Prompt Engineering techniques to standardize and ensure greater accuracy in the responses produced. 

\subsection{Applying RAG}
\label{sec:applicandoRAG}
One of the first decisions to be made is to choose the best strategy to segment the document, that is, how to perform the \textit{chunking} of the PDF files. A common \textit{chunking} strategy involves segmenting documents based on a specific number of \textit{tokens} and an overlap (\textit{overlap}). This is useful when dealing with sequential texts where it is important to maintain the continuity of the context between the \textit{chunks}.

\begin{figure*}[ht]
\centering
\includegraphics[width=\textwidth]
{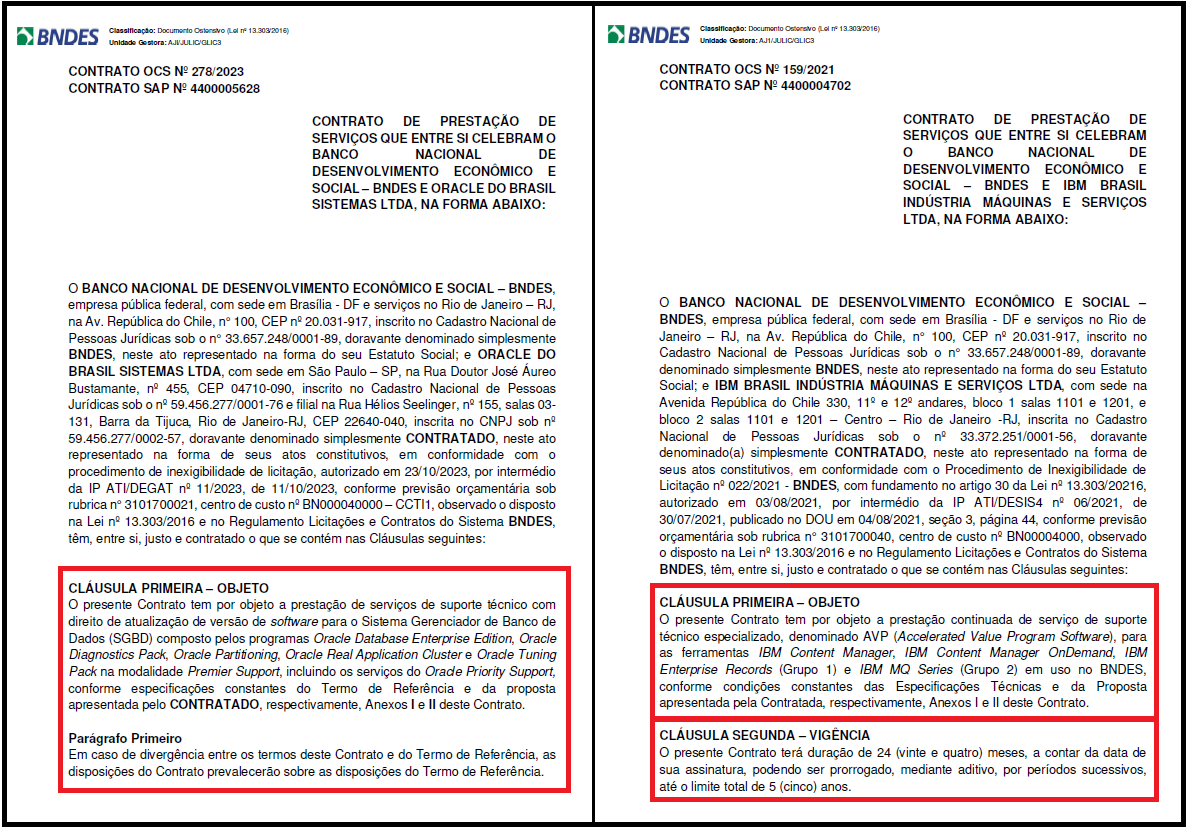}
\caption{Chunking applied to Contracts}
\label{fig:contracts}
\end{figure*}

Contracts have a standardized textual structure, organized into contractual sections. Therefore, sections with the same numbering or in the same vicinity describe the same contractual aspect, that is, they have similar semantics. For example, in the first section of contract documents, we always find the object of the contract. In this scenario, we can assume that the best \textit{chunking} strategy is to separate the \textit{chunks} by section of the document. In this case, the \textit{overlap} between the \textit{chunks} occurs by section, since the questions will be answered by information contained in the section itself or in previous or subsequent sections. For the contract page in the example in Figure~\ref{fig:contracts}, we would have a \textit{chunk} for the section on the object of the contract, another \textit{chunk} for the section on the term of the contract, that is, a \textit{chunk} for each clause of the contract and its surroundings. This approach ensures that each snippet represents a semantic unit, making retrievals more accurate and aligned with queries.

Having the contract section as the limit of the \textit{chunks} improves the relevance of the responses within a single contract. However, when increasing the number of contracts that the Contract\textit{360} intends to respond to, we observe the problem in correctly determining the contract to be treated. In the following example, we detail this aspect:

Consider the contract documents shown in Figure~\ref{fig:contracts}. showcases two service contracts between BNDES (Banco Nacional de Desenvolvimento Econômico e Social) and companies (Oracle do Brasil Sistemas Ltda. and IBM Brasil Indústria Máquinas e Serviços Ltda.), highlighting key clauses relevant to the provision of technical support and software updates. The contracts are presented in Portuguese, reflecting the original legal terms and specific obligations of each party. For instance, the contract with Oracle (Contract No. 278/2023) details the provision of services for Oracle Database and associated technologies, emphasizing software support and entitlement to updates. Similarly, the contract with IBM (Contract No. 159/2021) focuses on support services related to IBM Content Management software. The \texttt{"CLÁUSULA PRIMEIRA - OBJETO"} (first clause - object) details the object of the contract and a frequently asked question is: \textit{"What is the object of contract OCS 278/2023?"}. In this example, the RAG will store vectors containing the sections of both contracts, since this clause is common to both. However, when we inspect what is expressed in the \textit{chunk}, its content does not contain the contract number, Figure~\ref{fig:contracts}. Thus, with great probability, a query about a specific contract may return a segment (\textit{chunk}) unrelated to the contract, for example OCS 159/2021, being retrieved instead of the contract we want. In the case of our example, the \textit{chunk} referring to the question that should be returned is related to contract OCS 278/2023.

\begin{figure}[ht]
\centering
\includegraphics[height=.7\linewidth,width=1\linewidth]
{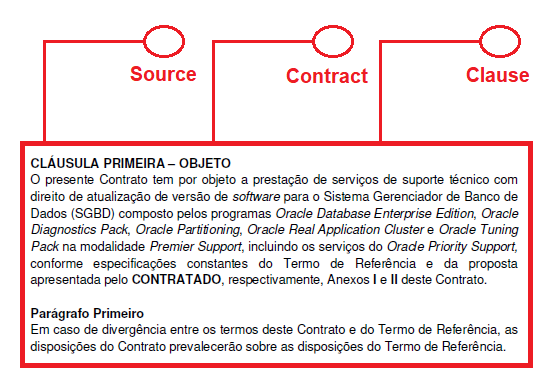}
\caption{Contracts metadata} 
\label{fig:metadados}
\end{figure}

To overcome this issue, it is necessary to add semantics to the \textit{chunks}, by including document metadata. And when accessing the \textit{vectorstore}, use this metadata to filter the information returned. In this way, we improve the relevance of the retrieved texts. Figure~\ref{fig:metadados} displays the most relevant metadata for the contracts (source, contract and clause). Where source is the name of the contract PDF file), contract is the OCS number and clause is the section title. Thus, for the question \textit{"What is the object of contract OCS 278/2023?"}, the \textit{chunks} of contract OCS 278/2023 are retrieved and then the similarity calculation is applied, retrieving the text segments to be sent to the LLM.

\subsection{Applying \textit{Text-to-SQL}}
\label{sec:applicandotext-SQL}
Contracts are dynamic and undergo several events like operational changes and management adjustments throughout their life-cycle. To deal with this complexity, organizations use contract monitoring systems, such as \textit{SAP Contract Life-cycle Management} and \textit{IBM Emptoris Contract Management}. These systems control several aspects, such as the technical person responsible for the contract, changes in the contractor's representative, and the end of the provision of services. During the contract term, these events can occur and significantly affect contract management.

The Contract\textit{360} retrieves those events from the Contract Management System (CMS) and incorporates them so the LLM can provide relevant responses to the user. Therefore, a \textit{text-to-SQL} technique was natural to implement the reasoning and decision-making task \cite{react2023} to obtain relevant responses from the CMS database to the contract managers.

The LangChain SQL Agent \cite{langchainSQLAgent} has proven to be a highly flexible tool for interacting with the CMS database. Upon system startup, our SQL agent establishes an authenticated connection to the database and retrieves the schema. When it receives a user question, it performs Entity Recognition, maps those entities to the database tables and columns, and prepares the SQL statement.

Ensuring the safety of our SQL agent is central. We validate each generated query to ensure it does not contain harmful commands, such as 'UPDATES,' 'DROP TABLE,' 'INSERT,' or any other command that can alter the database, providing a sense of security about the system's integrity.

Finally, the output generated from the executed SQL statement goes to a prompt generation stage for further analysis of the LLM.

\subsection{Applying Prompt Engineering}
\label{sec:applicandoPrompt}
The \textit{prompt} engineering technique provides a pattern for the style of responses and the reuse of the solution when accessing the LLM, as it provides instructions and context. Based on these observations, instructions were developed in the application to improve the responses. The instructions include basic guidelines, such as \textit{"Do not use prior knowledge"}, which ensures that the responses are based only on \textit{vectorstore} contracts, and specific instructions, such as \textit{"Whenever you answer a question about a contract, provide the OCS number."} Thus, the question \textit{"Do we have an Oracle Support contract?"} would have as a possible answer \textit{"Yes, we have an Oracle Database Support contract. The OCS number is 278/2023."}.

Maintenance and guidelines on how to use the chat context were also applied to ensure uniformity and coherence. For example, we inform the expected style for responses: \textit{"You should use a formal and objective tone."}, determining the style of LLM responses. Another guideline instructs LLM: \textit{"Given the chat history and the question asked, construct the response completely, without the user needing to review the history"}.

Finally, the context passed to the LLM can be useful to establish the style of the answers according to the role of the user of the Q\&A system. In the case of Contract\textit{360}, we have three roles: 1) contract manager; 2) contract management support; and 3) manager of the contract management support unit. For each of these roles a specific context was defined, for example for role 3 we have: \textit{"You are an assistant specialized in answering questions about administrative contracts, who provides management and summarized information about the contracts."}

With these three techniques we obtained more relevant answers. In the following section, we detail the implementation and the components used in the development of the system.

\begin{figure*}[ht]
\centering
\includegraphics[width=.7\textwidth]
{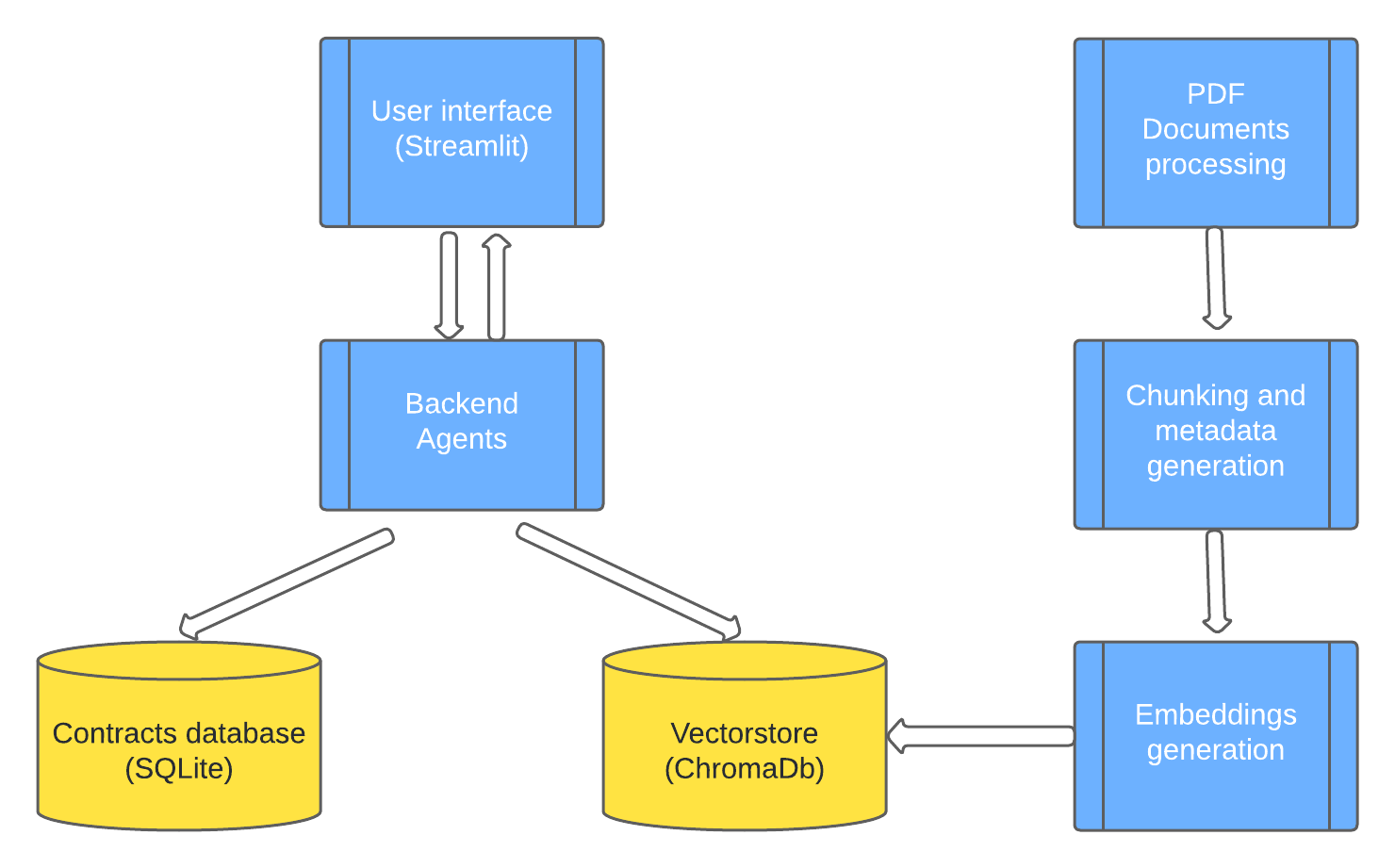}
\caption{Application architecture}
\label{fig:architecture}
\end{figure*}

\subsection{Applying Agents}
\label{sec:applicandoAgents}
In Contrato360, Agents play a pivotal role in orchestrating the flow of execution and enhancing the overall efficiency of the question-and-answer process. Also, considering the workflow on understanting the user query, an agent approach is a clever choice to implement this several specialized activities that needs to be taken in building the correct answer for the user. We designed three agents to implement this workflow.

As shown in figure \ref{fig:agents}, the Router Agent is central to its architecture, acting as the primary decision-making entity that orchestrates the flow of tasks needed to answer a user’s question. The "Router Agent" decides if the user's question is related to the Contract Manager domain, e.g., "How are you?", "Will Bologna FC win the 2025 Champions League?" or "Who is the contract manager for the Database support?". An out-of-topic question is redirected to the LLM with a context limiting its role to the domain of contract management. In A question on the contract domain will follow our workflow to find a relevant answer.

In the sequel, the Router Agent sends the user question to two specialized agents: a) SQL agent and b) RAG agent. The RAG agent retrieves from the \textit{vectorstore} chunks of documents similar to the user question. In parallel, a SQL agent retrieves form the CMS database content related to the user question. This architectural choice proved to be robust in the reports of the contract managers, as it semantically enriches the contract information, as shown in Figure \ref{fig:rag}.

One of the specialized agents in Contrato360 is the RAG (Retrieval Augmented Generation) Agent, responsible for retrieving relevant information from the contracts \textit{vectorstore}. When directed by the Router Agent, the RAG Agent searches for similar data chunks that can help contextualize the question. Another specialized component is the SQL Agent, which handles queries requiring structured data extraction from the contracts database. Upon receiving routing instructions from the Router Agent, the SQL Agent executes SQL queries to retrieve specific data points relevant to the user’s question. 

With all textual and information retrieve, another "Router Agent" craft an answer. If needed to add an visual information, the Graph Agent and LLM Answer Generation Agent add further depth to Contrato360’s response capabilities. The Graph Agent is tasked with creating visual representations, such as charts, when the Router Agent determines that a visual answer would better serve the user’s needs. This agent ensures that complex data can be conveyed in a clear and understandable format, enhancing user comprehension. Meanwhile, the LLM Answer Generation Agent works closely with the prompt generation module to produce coherent and contextually relevant textual responses. Together, these agents provide a multi-faceted approach to answering questions, combining data retrieval, visualization, and language generation to deliver comprehensive solutions.

\section{\uppercase{Architecture}}
The architecture of the Contrato360 application illustrates a comprehensive system designed to facilitate a question-answering application that integrates Large Language Models (LLMs), document processing, and databases. The architecture consists of three main layers: the User Interface Layer, the Backend Layer, and the Language Model Integration Layer, each playing its role in delivering accurate and context-aware responses to users.

The User Interface Layer is represented by the User Interface (Streamlit), which serves as the front-end of the application. This layer provides an interactive platform where users can input their queries and view the responses generated by the system. The interface directly communicates with the backend layer, sending user inputs for processing and displaying the responses generated by the various integrated components.

At the heart of the system lies the Backend Layer, which is primarily managed by the Backend Agents (Python and Langchain). This layer orchestrates interactions between the document processing, vector storage, contracts database, and the language model integration layer. The backend layer leverages Python and Langchain to handle the logic, task execution, and chat functionalities, particularly through OpenAI’s chat models. It processes user inputs received from the interface and interacts with both the Contracts Database and Vectorstore (ChromaDb) to retrieve relevant information necessary for formulating comprehensive answers.

Within the backend layer, the Contracts Database (SQLite) serves as the structured data source, storing structured information related to contracts. This component allows the system to handle contract-related questions by processing SQL queries generated by the backend agents. The contracts database responds to these queries with relevant data, which is then used to construct natural language responses for the user.

The Vectorstore (ChromaDb) is another vital component of the backend layer, acting as a storage solution for vectorized data, including document embeddings. It plays a key role in efficient similarity searches and retrieval tasks, enhancing the system’s ability to provide context-aware responses. The backend agents utilize the Vectorstore to match user queries against stored embeddings, enabling advanced semantic search capabilities. This component also stores embeddings generated from document processing, ensuring that data is readily available for future query matching.

The Language Model Integration Layer is responsible for transforming and embedding data for use within the system. This layer includes the PDF Documents Processing module, which ingests and preprocesses documents, particularly PDFs, to make them suitable for use within the application. This step involves reading and extracting text and relevant metadata, preparing the content for the next stages of processing. The Chunking and Metadata Generation component further refines the documents by dividing them into manageable chunks and generating metadata that improves retrieval efficiency, ensuring that the data is optimally split for better embedding generation and response times.

The final stage of the language model integration layer is the Embeddings Generation module, which converts the chunked documents and metadata into vector embeddings using LLM-based models like OpenAI Embeddings. These embeddings capture the semantic nuances of the text, facilitating efficient search and retrieval tasks within the system. Once generated, these embeddings are stored in the Vectorstore (ChromaDb), where they can be accessed for matching against user queries.

The overall workflow begins when a user inputs a question through the User Interface Layer, initiating a sequence of processes across the backend and language model integration layers. The backend agents handle query processing, interacting with the Contracts Database for SQL queries and performing semantic searches using embeddings from the Vectorstore. The document processing involves pre-processing PDFs, chunking the content, and generating embeddings that are then stored for efficient retrieval. The backend agents combine data retrieved from the contracts database and the Vectorstore to generate a coherent response, which is then presented back to the user through the User Interface Layer.

This architecture effectively combines the User Interface Layer, Backend Layer, and Language Model Integration Layer, enabling Contrato360 to function as a robust and powerful application for answering questions based on complex data sources. The seamless integration of multiple technologies ensures that users receive accurate and contextually relevant responses, enhancing the overall functionality and usability of the system.

\section{\uppercase{Evaluation}}
The experiment to validate the application was conducted by two IT contract specialists from BNDES. The system was prepared with 75 contracts (PDFs and data from the contract system). And to validate the relevance of the answers, \textit{benchmark} questions were prepared, from two distinct groups: "direct" and "indirect" questions. "Direct" questions are those that can be answered through the PDFs and their metadata. "Indirect" questions are those that obtain better relevance when searched in the contract system data. In Tables ~\ref{tab:direct} and ~\ref{tab:indirect} we present the users' perception of the quality of the answers. In the evaluation, the relevance of the answers was categorized as "Correct" and "Incomplete".\footnote{A third category would be "Incorrect", but this option was not obtained in any of the questions.}

We can observe that for the "direct" questions the system presents relevant answers for all experiments. However, in the "indirect" questions, despite being satisfactory, the results in one specific question were limited and incomplete. In our evaluation, these questions require a more elaborate semantic evaluation. In the first case, we realized that the concept of "Waiver of Bidding" was not well captured. We believe that an adjustment in the queries and/or in the prompt can add this type of semantics.

\begin{table}[ht]
\centering
\scriptsize
\begin{tabular}{|p{3.5cm}|c|c|}
\hline
\textbf{Question} & \textbf{Correct} & \textbf{Incomplete} \\
\hline
What is the subject of the OCS nnn/yy contract? & 10 & - \\
\hline
Do we have any contract whose subject is xxxx? & 9 & 1 \\
\hline
Do we have any contract with the supplier xxx? & 10 & - \\
\hline
Who is the manager of the OCS nnn/yy contract? & 10 & - \\
\hline
Who is the supplier of the nnn/yy contract? & 10 & - \\
\hline
What is the term of the OCS nnn/yy contract? & 10 & - \\
\hline
\end{tabular}
\caption{Direct Questions}
\label{tab:direct}
\end{table}

\vspace{0.1cm} % Adjust this value to control the vertical space between tables

\begin{table}[ht]
\centering
\scriptsize
\begin{tabular}{|p{4.0cm}|c|c|}
\hline
\textbf{Question} & \textbf{Correct} & \textbf{Incomplete} \\
\hline
How many active IT contracts do we currently have? & 10 & - \\
\hline
List the contracts that will end in the year yy? & 10 & - \\
\hline
How many contracts do we have with supplier xxxx? & 10 & - \\
\hline
How many contracts have we signed due to inflexibility? & 9 & 1 \\
\hline
How many DLs (Exemptions from Tenders) were contracted in yy? & - & 10 \\
\hline
Who are the managers of the contracts we have with company xxxx? & 8 & 2 \\
\hline
How many contracts does employee xxxx have under his/her management? & 8 & 2 \\
\hline
Show a summary of contract nnn/yy. & 10 & - \\
\hline
\end{tabular}
\caption{Indirect Questions}
\label{tab:indirect}
\end{table}

A key aspect observed from the users is the solution's capability to combine answers from both the structured data store and the contract's texts. This integration is perceived as a significant time-saving feature, as users typically need to locate the relevant contracts, open the respective PDFs, and manually search for additional information. The example below illustrates this. It identifies contract managers and outlines the penalties associated with contractual non-compliance. The system’s ability to deliver precise, context-relevant answers from contracts highlights its effectiveness in reducing manual search efforts for users. 

\begin{figure}[ht]
\centering
\fbox{\includegraphics[height=1\linewidth,width=.9\linewidth]{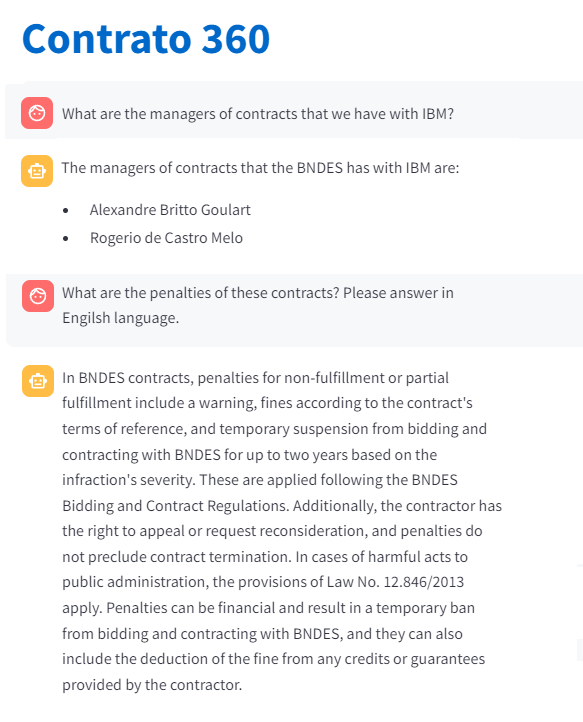}}
\caption{Contracts Q\&A Streamlit application} 
\label{fig:interface}
\end{figure}

In fact, by directly addressing questions with specific details, the system saves time and improves the user experience, as users can quickly access critical information without sifting through lengthy documents. Finally, the system's ability to automatically generate graphs using its Plotly agent, when a table of values is included in the response, has been positively received by users. This feature not only provides immediate visual insights, enhancing the understanding of the data, but also supports users in creating professional presentations. The integration of dynamic graph generation into the query response process significantly enriches the user experience, allowing for a more comprehensive analysis and efficient communication of contract-related information.

\begin{figure}[ht]
\centering
\fbox{\includegraphics[height=1\linewidth,width=.9\linewidth]{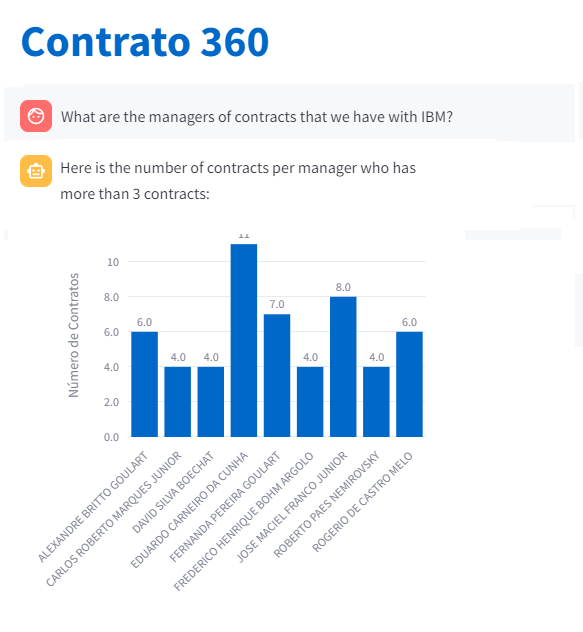}}
\caption{Plotly Agent} 
\label{fig:interface}
\end{figure}

\section{\uppercase{Conclusions}}
We developed a Q\&A application in the domain of service and product contracts, using PDF contracts and data from the Contract Management System as information sources. In this development, we employed four techniques to improve the relevance of the answers: 1) Augmented Retrieval (RAG) combined with semantic augmentation using metadata to retrieve information from PDFs; 2) Text-to-SQL, aggregating dynamic information from the contracts made available in the Contract Management System; 3) Prompt Engineering to contextualize, instruct and direct the answers produced by the LLM; and 4) Agents to call  the most appropriate approach depending on query context and determining the flow of execution of tasks in the system.

\begin{figure}[ht]
\centering
\fbox{\includegraphics[height=1\linewidth,width=.9\linewidth]{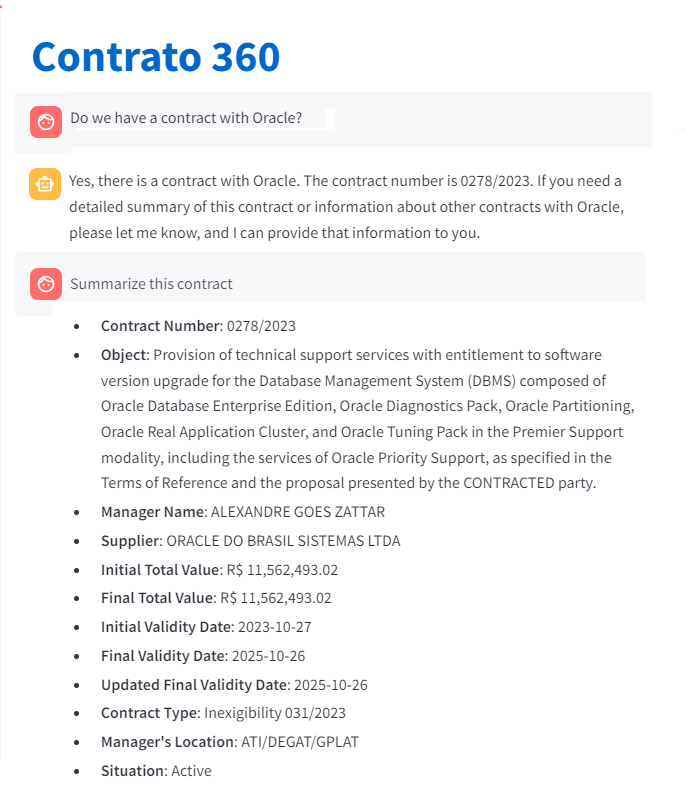}}
\caption{Contract Summarization} 
\label{fig:interface}
\end{figure}

The \ref{fig:interface} demonstrates the ability of Contrato360 in retrieving and summarizing contract information related to Oracle through a question-and-answer interface. When asked if there is a contract with Oracle, the system efficiently identifies the relevant contract, numbered 0278/2023, and provides a concise summary of its key details stored in the database. The summarized information includes the contract's object, which covers technical support and software upgrades for Oracle's Database Management System (DBMS), details about the contract manager, supplier, total value, validity dates, and the current situation. This functionality highlights the system's ability to streamline access to specific contract data, facilitating quick and accurate information retrieval for users by directly interacting with the database through natural language queries

In our experiment, we addressed an initial set of questions that were able to produce a robust system that meets current user needs. However, exploring other questions in depth will allow us to enrich the metadata and the set of queries that extract information from traditional systems.

Finally, to consolidate the techniques developed to address our application, we envision that building a system in a different problem domain may shed light on limitations and the possible need for refinement or adaptation. Such future exploration will not only reinforce confidence in the implementation of these techniques in real-world scenarios, but also pave the way for their optimization and possible customization for specific domains, ultimately increasing the utility and impact of LLMs in enterprise applications.

\bibliographystyle{apalike}
{\small
\bibliography{contrato360}}
\end{document}